# Airport Gate Assignment—A Hybrid Model and Implementation

## Chendong Li

Computer Science Department, Texas Tech University
2500 Broadway, Lubbock, Texas 79409 USA
chendong.li@ttu.edu

**Abstract**

With the rapid development of airlines, airports today become much busier and more complicated than previous days. During airlines daily operations, assigning the available gates to the arriving aircrafts based on the fixed schedule is a very important issue, which motivates researchers to study and solve Airport Gate Assignment Problems (AGAP) with all kinds of state-of-the-art combinatorial optimization techniques. In this paper, we study the AGAP and propose a novel hybrid mathematical model based on the method of constraint programming and 0 - 1 mixed-integer programming. With the objective to minimize the number of gate conflicts of any two adjacent aircrafts assigned to the same gate, we build a mathematical model with logical constraints and the binary constraints. For practical considerations, the potential objective of the model is also to minimize the number of gates that airlines must lease or purchase in order to run their business smoothly. We implement the model in the Optimization Programming Language (OPL) and carry out empirical studies with the data obtained from online timetable of Continental Airlines, Houston Gorge Bush Intercontinental Airport IAH, which demonstrate that our model can provide an efficient evaluation criteria for the airline companies to estimate the efficiency of their current gate assignments.

## Introduction

Aircrafts on the ground requires all kinds of diverse services, like reparation, maintenance and embarkation for passengers, that has to be guaranteed to be done within a very short time so that they must be in the right order. Growing flights congestion makes it necessary and compulsory to find ways to increase the airport operation efficiency. At this time research on airport gate assignment problem (AGAP) appears extremely significant on facilitating airlines to assess how many gates they should rent or purchase from airports to serve their own aircrafts (Lim & Wang 2005). Recently AGAP becomes one of the core components in the field of airport resource management and naturally appeals the close concentration of current researchers (Chun etc. 1999).

AGAP can be described as follows: Suppose an airline company owns the business of hosting a certain number of fights every day and in order to run the business smoothly it must purchase or lease a certain number of gates from an airport. The main mechanism of flight-to-gate assignments is that the airline companies can minimize the cost in the whole operational process. Efficient airport operations largely depend on how to gate aircrafts in a smooth flow of arriving and departing flights.

Different kinds of analytical models have been developed on gate assignment problem, such as Mangoubi and Mathaisel 1985, Vanderstraetan 1988, Cheng 1997, Haghani and Chen 1998. At the same time, various techniques have been applied to solve this problem. For instance, linear binary programming (Babic 1984), 0 – 1 linear programming (Bihr 1990), genetic algorithm (Gu & Chung 1999), mixed 0 – 1 quadratic integer programming and tabu search (Xu & Bailey 2001), multi-objective programming (Yan & Huo 2001), simulated annealing (Ding 2002), stochastic programming (Lim & Wang 2005). Most of these techniques are employed to minimize the passenger's walking distance.

The rest of the paper is organized as the follows. First, we formulate the problem integrating the techniques of both constraint programming and 0 – 1 mixed-integer programming with the objective to minimize the conflicts between any two adjacent aircrafts assigned to the identical gate. In the following, we describe the implementation issue and our experiments with the data from the online timetable of Continental Airlines, Houston IAH Airport under some specified assumptions. Moreover, we continue to interpret experimental results in detail, which further demonstrate the power and significance of our model. Finally, our conclusion is presented.

## Problem Formulation

We formulate the airport gate assignment problem as the constraint resource assignment problem where gates serve as the limited resources and aircrafts play the role of resource consumers.

The operation constraints consist of two items: 1) every aircraft must be assigned to one and only one gate. Namely, for a given gate it can be occupied by one and only flight at the same time. Also it can be free without holding any flight at curtain time. 2) For safety consideration, it is prohibited that any two aircrafts are assigned to the same gate simultaneously. In other words, if a gate is occupied by one aircraft, it cannot be assigned to another one until it has been released.

In fact, the airport gate assignment is a very complicated process; while for the sake of simplifying the problem, we mainly take into consideration of the following three factors:
• Number of flights of arriving and departure
• Number of gates available for the coming flight
• The flight arriving and departure time based on the fight schedule

## Notation

To model the gate assignment problem in the mathematical form, we first describe the following data sets and parameters.

Gates Set (Resources): $G = \{g_1, g_2, \cdots, g_c\}$ where c is the number of available gates;

Aircrafts Set (Consumers): $F = \{f_1, f_2, \cdots, f_n\}$ where n is the number of aircrafts.

For every aircraft $f_i$ ($1 \leq i \leq n$), we need the following notation:
• $a_i$: scheduled arriving time
• $d_i$: scheduled departure time
• $x_{i,k}$: decision variable
  $x_{i,k} = 1$ if and only if aircraft $f_i$ is assigned to gate $c_k$;
  $x_{i,k} = 0$ otherwise ($1 \leq i \leq n$, $1 \leq k \leq c$).
• $y_{i,j}$ : auxiliary variable
  $y_{i,j} = 1$ if $\exists k$, $x_{i,k} = x_{j,k} = 1$ ($1 \leq k \leq c$);
  $y_{i,j} = 0$ otherwise ($1 \leq i, j \leq n$).
• $b$: buffer time (constant)

Buffer time will lock the gate that has already been assigned to a certain aircraft before it arrives at the gate and after it leaves the gate. The goal of buffer time is to enlarge the interval between any two adjacent aircrafts assigned to the same gate, which will naturally decrease the probability of conflict between these two aircrafts.

We define *time interval* as the gap of gate locking time between two adjacent aircrafts $f_i$ and $f_j$ and the relation adjacent is defined as any aircrafts $f_i$ and $f_j$ that are assigned to the same gate consecutively. In other words, a certain gate is first occupied by $f_i$ and then sequentially by $f_j$, which indicates that there is no aircrafts assigned to this gate between $f_i$ and $f_j$. According the definition, the time interval locked for a particular aircraft (in terms of a given gate) equals to $[a_i-b, d_i+b]$.

A gate conflict (or conflict) is the scenario that it must lead to a collision between any two adjacent aircrafts because of the unreasonable gate assignment or caused by the real departure and arriving time, such as the delay of scheduled time. A gate conflict between any two aircrafts $f_i$ and $f_j$ if both of the following two conditions hold:
• Aircrafts $f_i$ and $f_j$ are assigned to the same gate, that is $y_{i,j} = 1$;
• There is an overlap between the two time intervals of two adjacent aircraft, that is
$[a_i-b, d_i+b] \cap [a_j-b, d_j+b] \neq \emptyset$, which is equivalent to $y_{i,k}*y_{j,k}(d_i-a_j)(d_j-a_i) \leq 0$ ($\forall 1 \leq i, j \leq n$, $i \neq j$, $\forall 1 \leq k \leq c$).

## Mathematical Model

With the objective to minimize the number of gate conflicts which depend on the gate assignment and the scheduled time, we use $p(i,j)$ defined as the probability distribution function on gate conflict between two aircrafts $f_i$ and $f_j$ if they are assigned to the same gate. Then the gate assignment model can be formulated as follows:

$$Min \sum_{i \in N} \sum_{j<i, j \in N} y_{i,j} * E(p(i,j)) \quad (1)$$

Subject to

$$\sum_{i \in N} \sum_{k \in C} x_{i,k} = 1 \quad (2)$$

$$\sum_{i \in N} \sum_{j<i, j \in N} \sum_{k \in C} (x_{i,k} * x_{j,k}) = y_{i,j} \quad (3)$$

$$y_{i,k} * y_{j,k} * (d_i - a_j) * (d_j - a_i) \leq 0 \quad (4)$$

$$x_{i,k} \in \{0,1\} \quad (5)$$

$$\forall 1 \leq i, j \leq n, i \neq j, \forall 1 \leq k \leq c \quad (6)$$

In the model, N is the integer set of the numbers of all the flights needed to be assigned to gates and C stands for the set consisting of all the gates available to host flights. $f_i$ and $f_j$ means different flights and c is the number of gates.

Equation (1) denotes the objective function, which could be simplified by applying the uniform distribution. Specifically, in this model, $E(p(i, j))$ equals to $1/(a_i-d_j+2b)$. In other word, we applied the following equation to our current objective function (1).

$$E(p(i,j)) = \frac{1}{a_i - d_j + 2b} \quad (7)$$

Equation (2) indicates that each aircraft is assigned to one and only one gate. Equation (3) represents the numerical relationship between $x_{i,k}$ and $y_{i,j}$, which presents a method to compute the auxiliary variable $y_{i,j}$ from $x_{i,k}$. Equation (4) guarantees that one gate can only be assigned to one and only one aircraft at the same time. The scenario that there is an overlap between any two adjacent aircrafts is stated in this constraint. Some additional constraints in the real operations, such as that some particular aircrafts should be assigned to certain gates (like VIP gates), are out of consideration in the current formulated model. However, we can easily add such preference constraints to the model. Equation (5) represents the decision variables, of which the value is binary.

The proposed model can provide an efficient evaluating criteria for airlines to evaluate their current gate assignment, as the model has indicated the potential objective to minimize the number of gates. For example, if airline authorities want to evaluate the efficiency of the gate assignment of certain number of flights (published as timetable or schedule for passengers' reference) at a certain airport, they can calculate the value of the objective function in our proposed model based on the published schedule. Intuitively once the value is big, such as bigger than 10, it indicates that the current gate assignment is not good and the authority should consider the reassignment or modify current flight schedule. However, if the value is quite small, such as very near to 0, it denotes that the current gate assignment is almost optimal in the scenario that the number of available gate is fixed at present.

## Implementation

In the Optimization Programming Language we encode our model into OPLscript as shown in Figure 1 and run the program in ILOG OPL studio 3.7.1 IDE. In the OPLscript of Figure 1, arrtm, dptm, nbFlt, and nbGate stand for arriving time, departure time, number of Flight and number of Gate, respectively.

We run our program on Dell server PE 1850 under the configuration of Intel(R)Xeon(TM) CPU 3.20GHz, 3.19GHz, 2.00G of RAM.

```
int nbFlt = ...;
int nbGate = ...;
range Gate 1 .. nbGate;
range Flt 1 .. nbFlt;
var
   int assign[Flt,Gate] in 0..1,
   int y[Flt,Flt] in 0..1;
float+ arrtm[Flt] = ...;
float+ dptm[Flt] = ...;
minimize
     sum (i in Flt, j in Flt : arrtm[j] - dptm[i] >0 ) y[i,j] / (arrtm[j] - dptm[i])
subject to {
        forall(i in Flt, j in Flt : j<i )
                   sum(k in Gate)assign[i,k] * assign[j,k] = y[i,j];
        forall(i in Flt)
                   sum(k in  Gate) assign[i,k] = 1;
        forall(i in Flt, j in Flt, k in Gate : j<i)
                   y[i,k]*y[j,k]*(arrtm[i] - dptm[j])*(arrtm[j] - dptm[i]) <= 0;
        };
```

Figure.1 Assignment.mod

## Experiment

In this part, we will describe how we conduct all the experiments and report relevant results. Before starting our formal experiment we first obtain the raw data and analyze the data especially due to the large data size. In the following steps, we run the program and collect the experimental results. At the end of this part we refer to our future research directions to improve the experiment.

**Data Analysis**

First we consider an airport with three gates and a schedule of six aircrafts first. To apply the proposed model, we first calculate the matrix of $E(p(i, j))$ (we use the common accepted buffer time as b = 15 for illustration).

In term of large date size, we choose the online timetable of Continental Airlines as our raw data. Based on the timetable of Continental Airlines, we extract the departure time of all the flights in a whole day, from 6:00 A.M. to 23:59 P.M., leaving from Houston George Bush International Airport -IAH.

From the schedule, we obtain the departure time of 996 flights in all, in which we did not separate the real timetable that might be different from day to day. For instance, a certain fight may fly to a given destination on special days in a week, like Monday, Wednesday and Friday. In our experiment, we add all the flights with different flight number to out flight set, which will be considered for the gate assignment. Also, we assume that all the flights will leave the airport an hour later after their arrivals, i.e. every flight will stay in the airport for one hour.

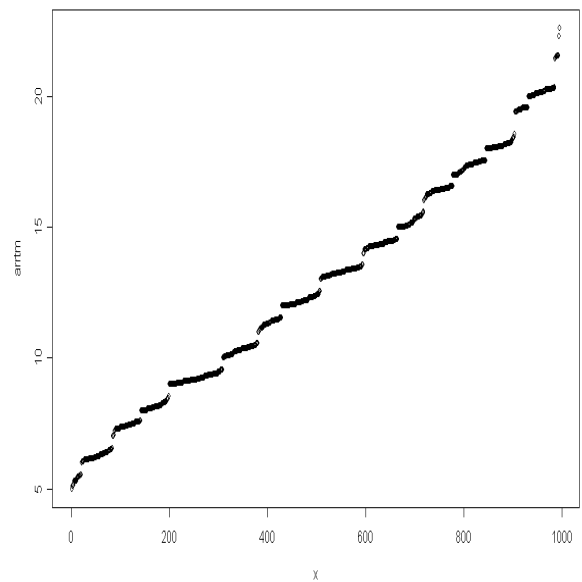

Figure 2 996 Flights scatter plot diagram

The total number of the flights to be considered in the experiment is 996. We plot 996 flights into the coordinate system with arrtm (abbreviated for arriving time) as vertical axes and x (denoting the number of flights) as horizontal axes, as illustrated in Figure 2. Each small circle in the scatter plot diagram stands for one single flight.

Experiment 1: Using small data set to make sure the model can get the right output.

Experiment 2: Medium data set with distinct number of gates to compare the running time and objective value. In experiment 2, we assign 33 flights with different number of gates ranging from 1 to 10 and finally 50 as shown in table 1.

Table 1 33 flights assigned to different nuber of gates

| Number of Gates | Running Time(sec.) | Value of Objective Function |
| --- | --- | --- |
| 1 | 0.03 | 878.0319 |
| 2 | 153.85 | 169.4055 |
| 3 | 311.75 | 28.7304 |
| 4 | 874.01 | 3.5644 |
| 5 | 3.895 | 0.000 |
| 6 | 4.53 | 0.000 |
| 7 | 5.19 | 0.000 |
| 8 | 5.38 | 0.000 |
| 9 | 5.66 | 0.000 |
| 10 | 6.17 | 0.000 |
| 15 | 7.73 | 0.000 |
| 20 | 9.61 | 0.000 |
| 30 | 13.86 | 0.000 |
| 50 | 23.19 | 0.000 |

*Note: The Running time presents the average running time.*

Experiment 3: Large data set to compare the running time and objective value. We assign 996 flights to 70 gates and 65 gates respectively.

**Experimental Results**

In experiment 1 where we deal with small data set, the optimal solution with objective value is 287.0787 indicates that the gate conflicts are inevitable because only 3 gates available. When we enlarge the available gate number to 6, the gate conflicts decrease dramatically and reach the objective value smaller than 3.8615, which is much better compared to 287.0787 conflicts with 3 gates.

Table 1 presents the running time and the value of objective function due to different numbers of gates available as in experiment 2. From this table we can see clearly that when the number of gates equal to 4, the running time is as high as 874.01s, which is the maximum running time of the whole experiment 2. Moreover, it indicates that when the gate number is 5 it is ideal to avoid all the gate conflicts. In fact, airlines may choose 4 gates considering the large cost of leasing or buying a new gate.

From experiment 2, we can see that if airline authorities attempt to assign a certain number of flights to different number of gates, they can easily compare their current assignments with other assignments based on distinct number of gates available. Specifically, suppose that an airline's current assignments make use of *n* gates to run its business and the authorities find that it is feasible to assign the current flights to *n - 1* gates with the same objective value 0.000 (no gates conflicts), the airline probably takes action to reduce the redundant gate and revise its current assignments so that it can minimize the cost of running the business because the current gate assignments are not optimal. This can indicate that our proposed model could provide a very powerful tool for airline companies to estimate the efficiency of their current gate assignment.

**Future Work**

Optimizing the objective function with the conditional probability and the fuzzy theory is a potential direction.

We attempt to build precise evaluation criteria with the ability to deal with aircraft-to-gate assignment and handle the uncertainty in the real life airport daily operations, i.e. it is a very frequent phenomenon that aircrafts always arrive later than the scheduled time because of some uncontrollable factors like the weather condition. We plan improve our model to search the most robust airport gate assignment or second most robust airport gate assignment (considering the time expense) accurately and effectively.

Limit the number of conflicts in the objective function, like to a constant M (we can set M=0, which means no conflicts or other practical value) and modify the current schedule is also a practical and significant direction.

**Conclusion**

During the airline daily operations, assigning the available gates to the arriving aircrafts based on the published schedule is a very important issue. In this paper, we propose a hybrid model by employing the techniques of constraint programming and 0 - 1 mixed-integer programming. Our model is not only simple, easy to modify, but also pragmatic, feasible and sound. The designed experiments demonstrate that the proposed model is of great significance to help airline companies to estimate their current gate assignments efficiently and correspondingly guide the airlines to take actions to minimize the cost of leasing or purchasing the unnecessary gates and reach the optimal status in which they keep running their business smoothly. Although the experimental data is from a specific airport, the model has the generality that can be applied to other airports. Furthermore, the model allows the airline authorities to add logical constraints and preference constraints freely based on the distinct requirements of a particular airport.

**Acknowledgement**

I would like to express my gratitude to Dr. Yuanlin Zhang for his discussion and support during the research.